\def\year{2019}\relax
\documentclass[letterpaper]{article} %DO NOT CHANGE THIS
\usepackage{aaai19}  %Required
\usepackage{times}  %Required
\usepackage{helvet}  %Required
\usepackage{courier}  %Required
\usepackage{url}  %Required
\usepackage{graphicx}  %Required

\begin{document}
\title{Clustering Player Strategies from Variable-Length Game Logs in \emph{Dominion}}
\author{Henry Bendekgey\\
Pomona College\\
Claremont, CA\\
henry.bendekgey@pomona.edu}
\maketitle
\begin{abstract}
We present a method for encoding game logs as numeric features in the card game \emph{Dominion}. We then run the manifold learning algorithm t-SNE on these encodings to visualize the landscape of player strategies. By quantifying game states as the \emph{relative} prevalence of cards in a player's deck, we create visualizations that capture qualitative differences in player strategies. Different ways of deviating from the starting game state appear as different rays in the visualization, giving it an intuitive explanation. This is a promising new direction for understanding player strategies across games that vary in length.
\end{abstract}

\section{Introduction}

The study of artificial intelligence for games has historically been concerned with understanding strategy. However, until recently, the focus has been on the ideal strategy, and designing an agent that can outperform or compete with the best humans. As a result, the AI community has made significant advances in developing agents with a similar skill level to humans, even if their behavior is not particularly similar \cite{balance}.

These high performance agents reflect big steps forward for the AI community, but offer little to the game development community. A game developer is less likely to be interested in a perfect agent, and more likely to be concerned with designing an AI that is fun to play against, or models a specific personality. Here, we consider personality to be a tendency towards a specific playstyle or strategy.

Personality-based agents also have significant implications for game design. A well-designed game should ideally be robust to a variety of player strategies, but it can be difficult to understand the landscape of those strategies in the game development process. Personality-based agents can automate that process, allowing game designers to understand how individual mechanics or rule changes affect the relative strength of various game-playing approaches.

More broadly, games provide a closed environment for understanding the landscape of individual preferences. Ultimately, we hope to extract personalities or strategies that can be applied to a new task or set of game mechanics, and model how that player would behave.

The natural question that follows from this is how we learn these strategies. Ideally, we could develop these strategies from only the rules of the game. In reality, this process is significantly easier for an already-released game for which we can observe properties about the meta-game, and how players interact with it. This field of work's application to game balance can still apply in the case of already-released games---many online games are constantly updated to maintain balance, and other games, like \emph{Dominion}, are concerned with designing well-balanced expansions.

In this work, we examine the popular strategy card game \emph{Dominion}, and propose a method for encoding games traces into numeric features. After encoding individual players' gameplay, we use the t-SNE dimensionality reduction technique to visualize the landscape of player strategies. We see promising results that reflect different ways of diverging from the starting game state.

We explore game strategy as a proxy for player personality; once a field of strategies is observed, they can be clustered into player personalities based on our belief about a player's propensity to choose one strategy over another.

\section{Background}

\subsection{Dominion}
\emph{Dominion} is a well-studied game by the game AI community \cite{dominionValFuncs,newCards,personality}. It is a card game in which players aim to build a deck of high-quality cards starting from a seed of 10 low-quality cards. Cards are drawn from a player's deck and then used to buy new cards from the common pool, which are cycled into the player's deck for future turns. 

The game consists of three primary card types: action cards, which allow a player to execute specific abilities; treasure cards, which count as currency for the player to be able to buy new cards at the end of her turn; and victory cards, which are worth points at the end of the game but useless in hand. The diversity in card mechanics leads to variability in player strategy. 

\emph{Dominion} exhibits many properties that make it a difficult game to extract knowledge from or design AI agents for. Firstly, it is stochastic, with players drawing from a shuffled deck every turn. Thus player decisions at time-step $t$ cannot be directly compared.

Further, the set of cards in the common pool from which players build their deck changes from game to game. The base game of \emph{Dominion} consists of 26 cards of which 10 are chosen for each game. These are added to a pool of 7 universal cards that are used in every game. There are approximately five million ways to set up the game from the base game alone, and with 359 total cards in all expansions, the combinatorial explosion means that most card sets have never been observed before. 

This poses an interesting opportunity to game AI experts: \emph{Dominion}'s mechanic space is changeable, and by developing theories on a single set of cards, we can see how they hold up to mechanic changes. In the context of this work, we examine the strategies present under a single set of cards, with the goal of extrapolating these strategies into player personality models which can be applicable to a different set of cards.

\subsection{Related Work}

Previous work on player persona modeling has largely been motivated by the development of utility functions that reflect players' propensities towards certain actions \cite{Holmgard2014}. It has been shown that these utility functions can be developed with evolutionary computation and combined with the Monte-Carlo Tree Search algorithm to develop agents that model certain personalities \cite{Holmgard2018}. However, the motivation behind these papers is to examine and encode intuitive player desires, like ``finish as quickly as possible" or ``defeat all the enemies." In a game where it is harder to identify underlying motivations, these models become less applicable.

In an earlier work on clustering of \emph{Dominion} personas, Kevin Gold developed intuitions for what certain players strive for, and then used Bayesian network models to cluster players into groups \cite{personality}. However, the final clusters did not reflect the priors, and the vast majority of players were moved into a single cluster. Gold concluded that clusters may not as discrete as we might want. Many players will try a combination of synergies in a single game, just as many players enjoy both speedy gameplay and defeating monsters, and so in building off of Gold's conclusions we expect a continuum between different ways of playing. 

Gold's paper proposed two specific models for predicting card buys: a trigram model based on two previous buys, and a naive Bayes model, based on all cards currently in the player's deck. The author notes that the ``natural follow-up experiment should be to determine whether EM can assign players to one model structure or other based on how well the models capture player behavior" \cite{personality}. Thus Gold proposed clustering players on which factors inform their decision making.

However, the issue of examining what informs a player decision is particularly difficult in \emph{Dominion}. Because victory cards are useless in the deck until the end of the game, general strategies involve so-called \emph{engine building}: at the start, a player buys high quality cards so that towards the end she can buy as many victory cards as she can. However, when this switch occurs, and to what degree it is a binary switch as opposed to a gradual one, remains an open question.

Games can vary greatly in length, and with multiple possible end conditions, it can be difficult to estimate what stage a game is in. A recent work found that a neural network achieved better performance at \emph{Dominion} when replaced with two neural networks for the early and late game, and evolving the change point between them \cite{dominionValFuncs}.

\subsection{Strategy}

These papers are primarily concerned with modeling a player's decision-making process in order to understand her personality. This can be done by considering a player's actions across multiple games. In contrast, in this paper we are concerned with understanding a player's strategy, the sequence of actions taken in a single game. 

Previous work on knowledge extraction from games has considered two methodologies for encoding play traces: either as a sequence of decisions \cite{Osborn2017} or a sequence of game states \cite{andersen2010gameplay,liu2011feature,moment}. Because of the stochastic nature of \emph{Dominion}, sequences of decisions are not directly comparable, so we will primarily encode the traces as sequences of game states.

\section{Methodology}

\begin{figure}
\begin{center}
\includegraphics[width=\linewidth]{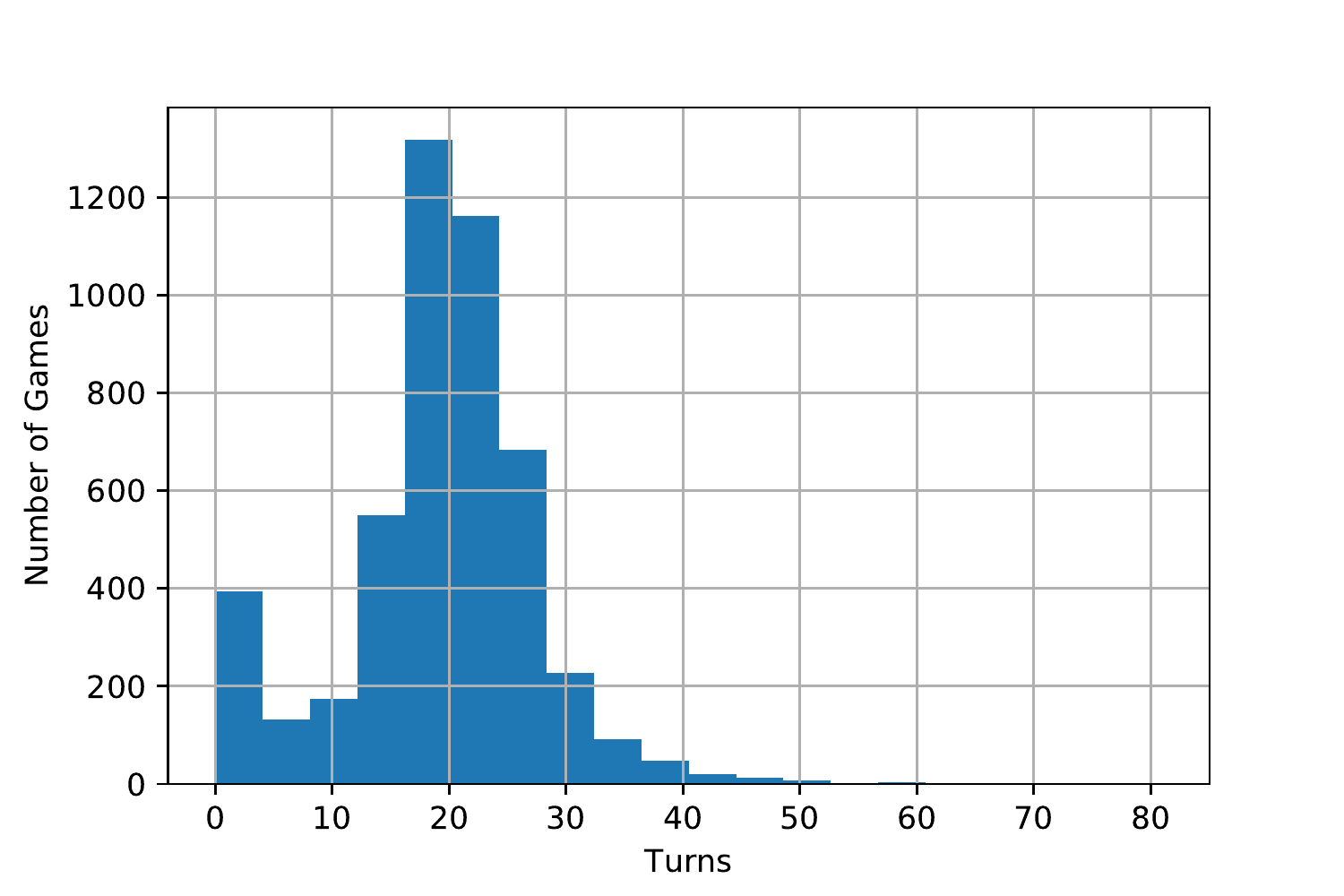}
\end{center}
\caption{Histogram of game lengths, measured in turns, for all player counts.}
\label{turndist}
\end{figure}

\begin{figure*}
\begin{center}
	\includegraphics[width=0.45\textwidth]{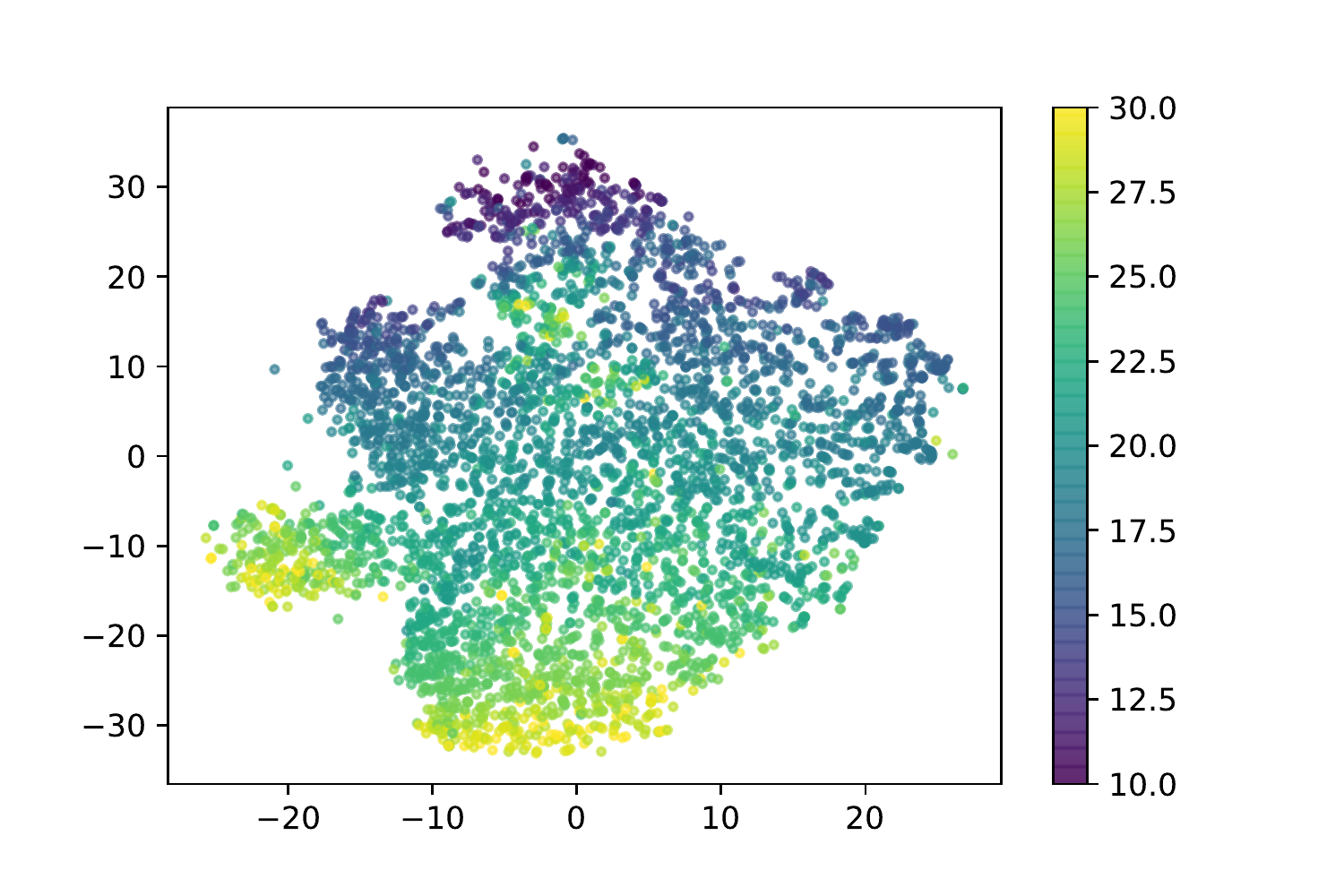}
	\includegraphics[width=0.45\textwidth]{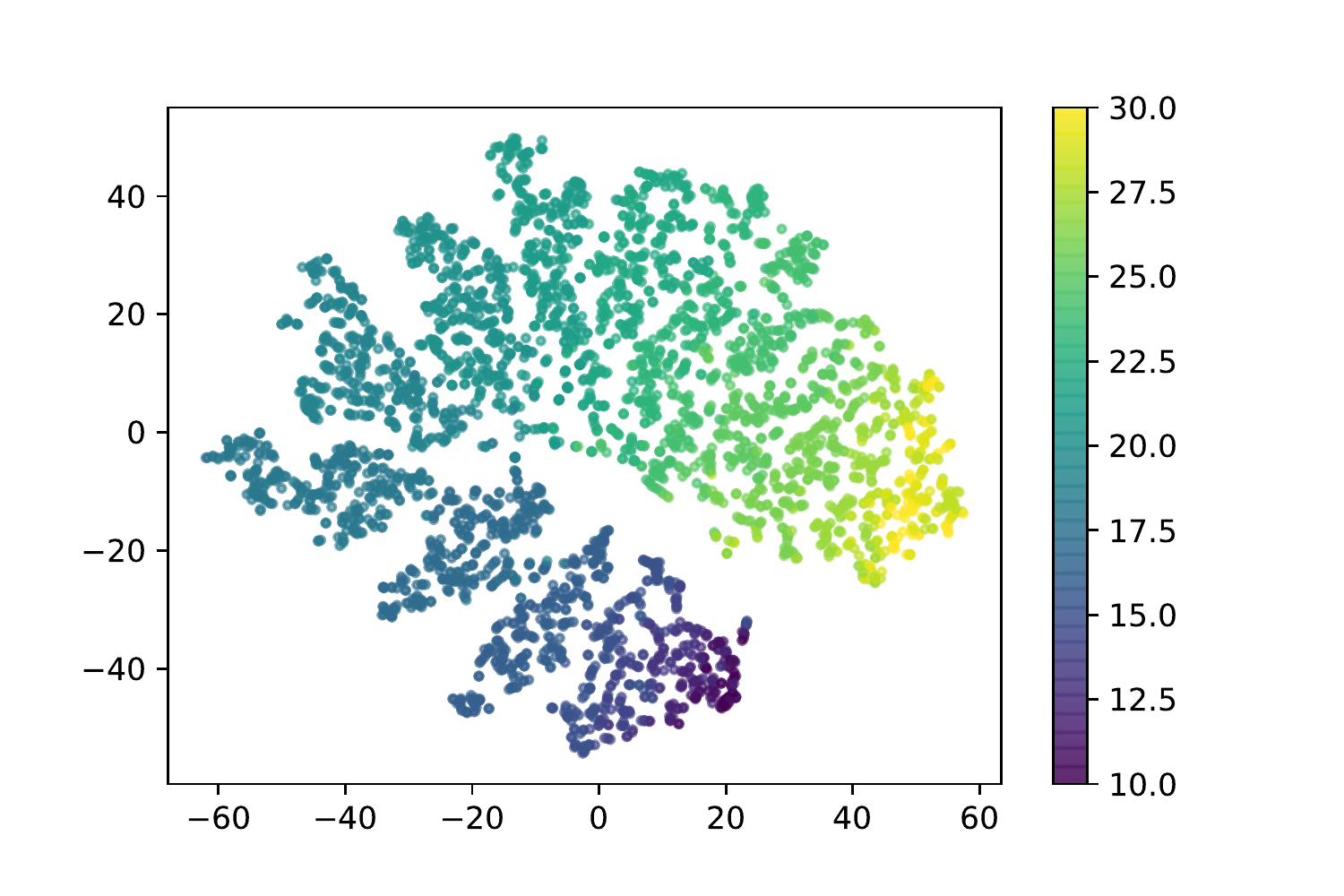}
\caption{2-dimensional embeddings of player traces, with color encoding the length of the trace in turns. On the left is the turn encoding; on the right, the game state encoding.}
\label{cl12}
\end{center}
\end{figure*}

\subsection{Data}
For this study we make use of data provided by dominion.isotropic.org, a server created by fans for online play of \emph{Dominion}. In 2013, the server was shut down due to the official licensed implementation of online \emph{Dominion} going live, but all of the game logs are still available online. We examine logs from June 2011 to March 2013, totaling almost eleven million files. 

Despite there being eleven million total logs, the most popular card-set appears in only 3,012 games. The 10 cards were: \emph{Cellar, Market, Militia, Mine, Moat, Remodel, Smithy, Village, Woodcutter,} and \emph{Workshop}. For this study, we want to examine strategies under a closed set of rules, so we use this as our data. We are considering each player's game trace individually, so each player in a multiplayer game will have her own encoding.

Because some of the traces are 1-player games where the player appears to be experimenting with card combinations or playing illogically, we subset our data to only 2+ player games of length 10-30 turns. We found this to be the range of ``normal" game lengths, see Figure \ref{turndist}. We are left with 2,795 traces.

\subsection{Feature Encoding}

We propose a new encoding of game traces and compare its results with three naive encodings. For the first naive encoding, we create one feature for the number of each card played on each turn, and another feature for the number of that card bought on that turn. Given that there are 17 cards and 30 turns, the data is 1,020-dimensional. We will refer to this as the turn encoding method.

Next we try clustering on game state space, where each turn, we encode the number of cards of each type present in the player's deck at the end of the turn. In this case, the feature space is 510-dimensional. Further, we let the last game state hang, meaning that for a game that ends on turn 20, the feature for turns 20-30 are all identical, and reflect the end-game deck composition of that player. We call this the game state encoding method.

For the third encoding, we use the same methodology as the second encoding, but only encode the first 10 turns of the game and ignore everything that happens after that. The feature space here is 170-dimensional. We call this the opening encoding method.

Finally, for our proposed encoding, instead of the features representing the \emph{amount} of each card in a player's deck, we encode them as the \emph{proportion} of each card in the player's deck. Just as in the game state encoding method, we let the final state hang. Here the feature space is 510-dimensional. We will refer to this as the normalized encoding method.

\subsection{Manifold Learning}

We run t-distributed Stochastic Neighbor Embedding (t-SNE) on these features to visualize the topology of the point cloud. t-SNE is a manifold learning technique that allows for the detection and representation of non-linear structure in the high-dimensional data, keeping nearby points close together in the resulting low-dimensional embedding \cite{tsne}. In the case of \emph{Dominion} play traces, we encode the games into a set of features and use t-SNE to plot these games such that traces that are close to each other on the plot are similar in their features. Because \emph{Dominion} has some well-known strategies, we can verify t-SNE's output by looking for regions corresponding to these strategies. 

The axes of the resulting plot have no direct interpretation. We use the Barnes-Hut approximation \cite{van2013barnes} to improve running time and compute the embedding efficiently.

\section{Results}

In the first two encodings, the number of turns in the trace dominated the clustering results (see Figure \ref{cl12}). In the turn encoding, the sparsity of features for short games causes them to all be close to one another, resulting in the above figure. In the game state encoding, the divergence of the magnitude of the game state vectors has a similar effect. This is consistent with other methods that have been used to visualize game traces \cite{Osborn2017}. However, because a player has limited control over the game length, it seems inappropriate as the defining characteristic of the play trace.

\begin{figure}
\begin{center}
	\includegraphics[width=\linewidth]{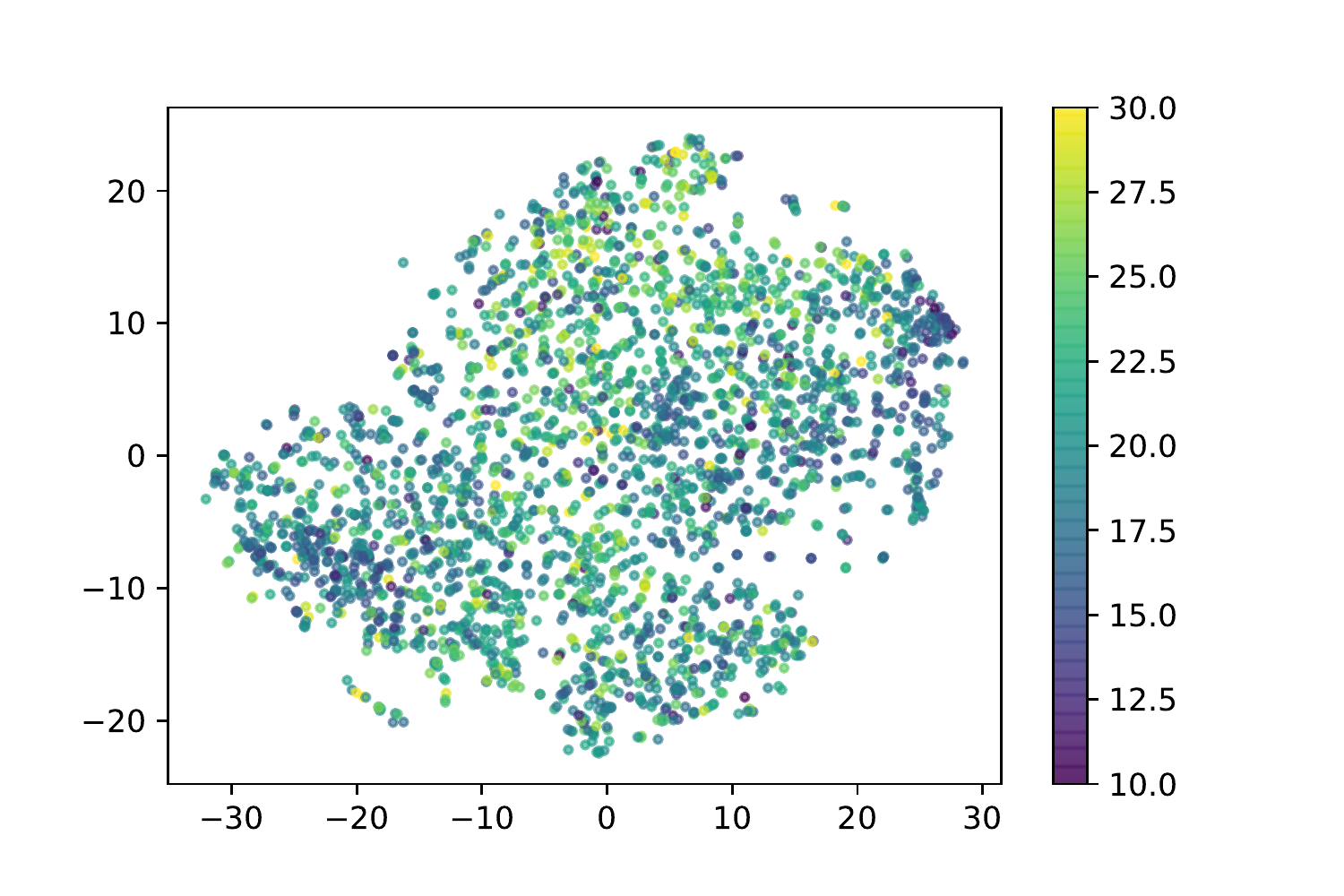}
\caption{t-SNE embedding of features in opening encoding method, where only the first 10 game states are used, with color representing the length of the trace in turns.}
\label{first10}
\end{center}
\end{figure}

In the opening encoding, we clustered on only the first 10 turns. Because we are only observing games of length 10-30 turns, this solves the problems observed in the first two clusterings, and we get good mixing of game length (see Figure \ref{first10}). However, we are leaving out important information. We are clustering not on player strategies but on player opening strategies.

\begin{figure}
\begin{center}
	\includegraphics[width=\linewidth]{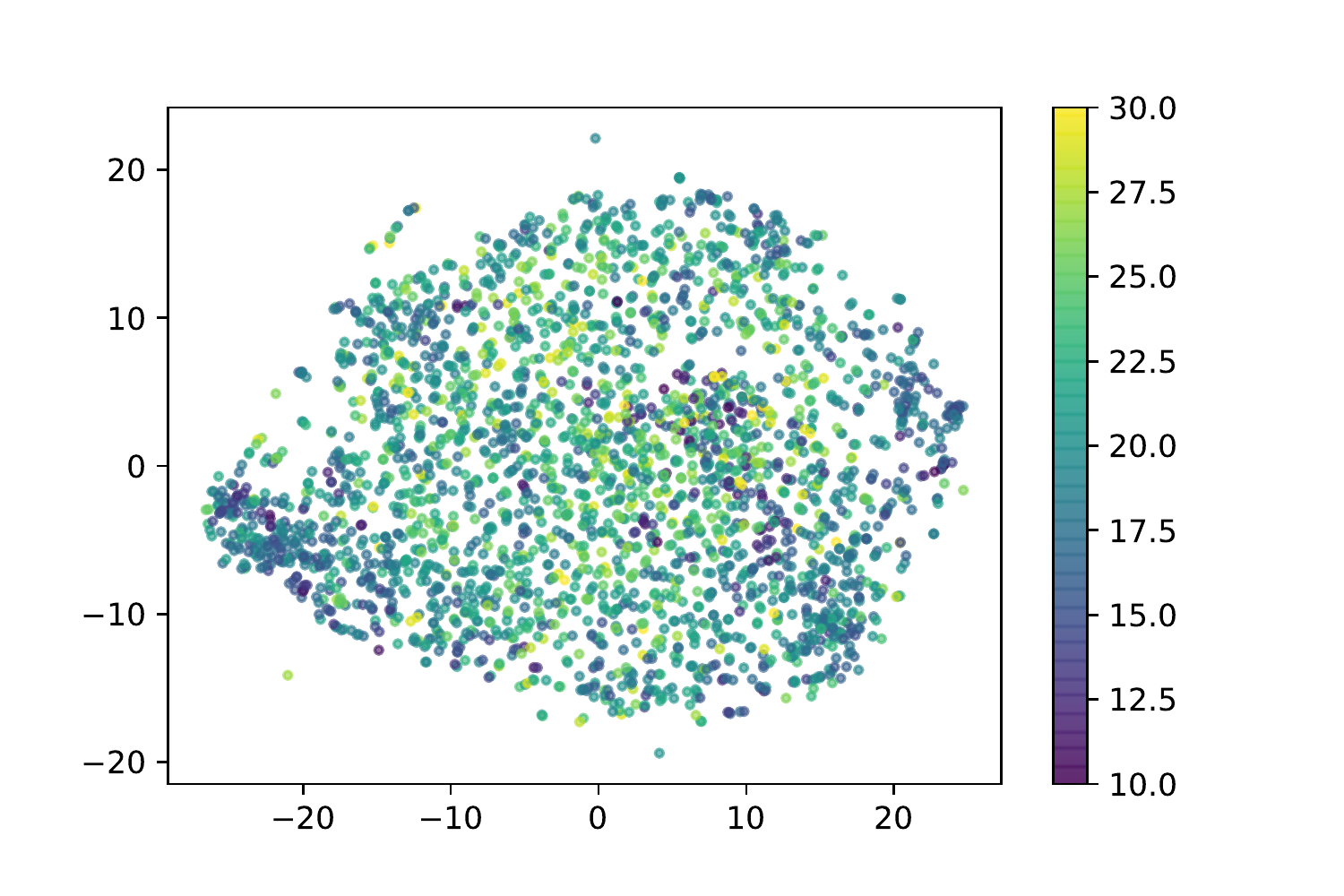}
\caption{t-SNE embedding of features using the normalized encoding method, with color representing the length of the trace in turns.}
\label{prop}
\end{center}
\end{figure}

\begin{figure}
\begin{center}
	\includegraphics[width=\linewidth]{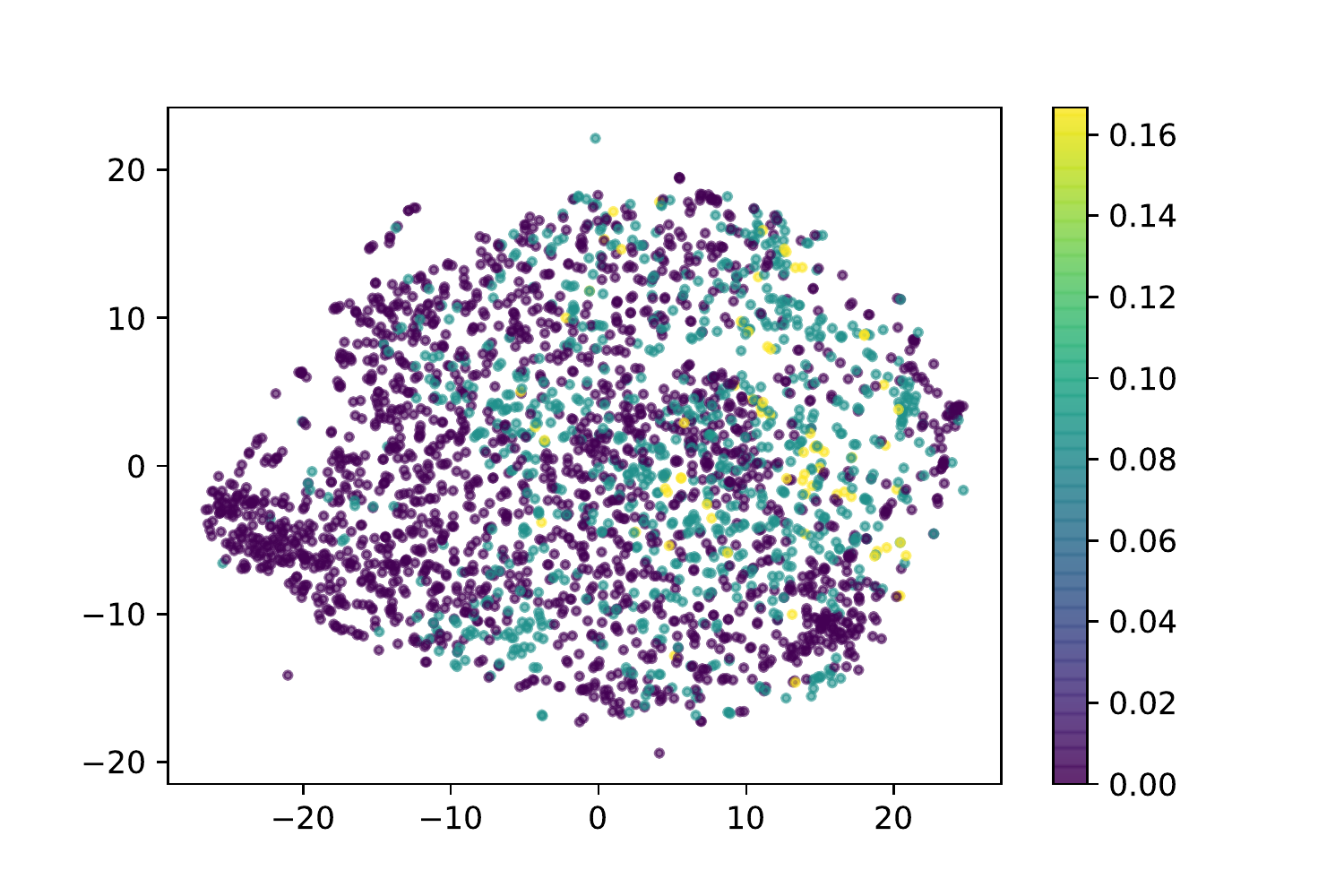}
\caption{Normalized feature embedding, with color representing the proportion of \emph{Villages} in the player's deck at the end of turn 2.}
\label{village}
\end{center}
\end{figure}

Finally, we consider the normalized encoding results. Because cards are being drawn randomly from the player's deck, there is little difference between two decks of the same proportional composition but different sizes. One substantive difference is the rate of intake for new cards, because a smaller deck will more quickly draw a card recently bought than a bigger deck. The other difference is that when it comes to victory cards we care about the number owned, not the proportion, because the sum of victory card values decides who wins.

In exchange for these simplifications, we end up with something more agnostic to game length. Intuitively, we are encoding the game state as the relative prevalence of game features instead of the absolute prevalence. Geometrically, this is forcing each game state to live on the $l_1$ unit ball, and encoding a game trace as a movement across the surface of that ball starting from the common seed deck. Regularization is a common data science tool which we can leverage here to compare varied-length games.

By encoding the deck as proportions, game state vectors are not diverging in magnitude, and thus longer games only appear distant from short games if they are qualitatively different, too.

\begin{figure*}
\begin{center}
	\includegraphics[width=.45\textwidth]{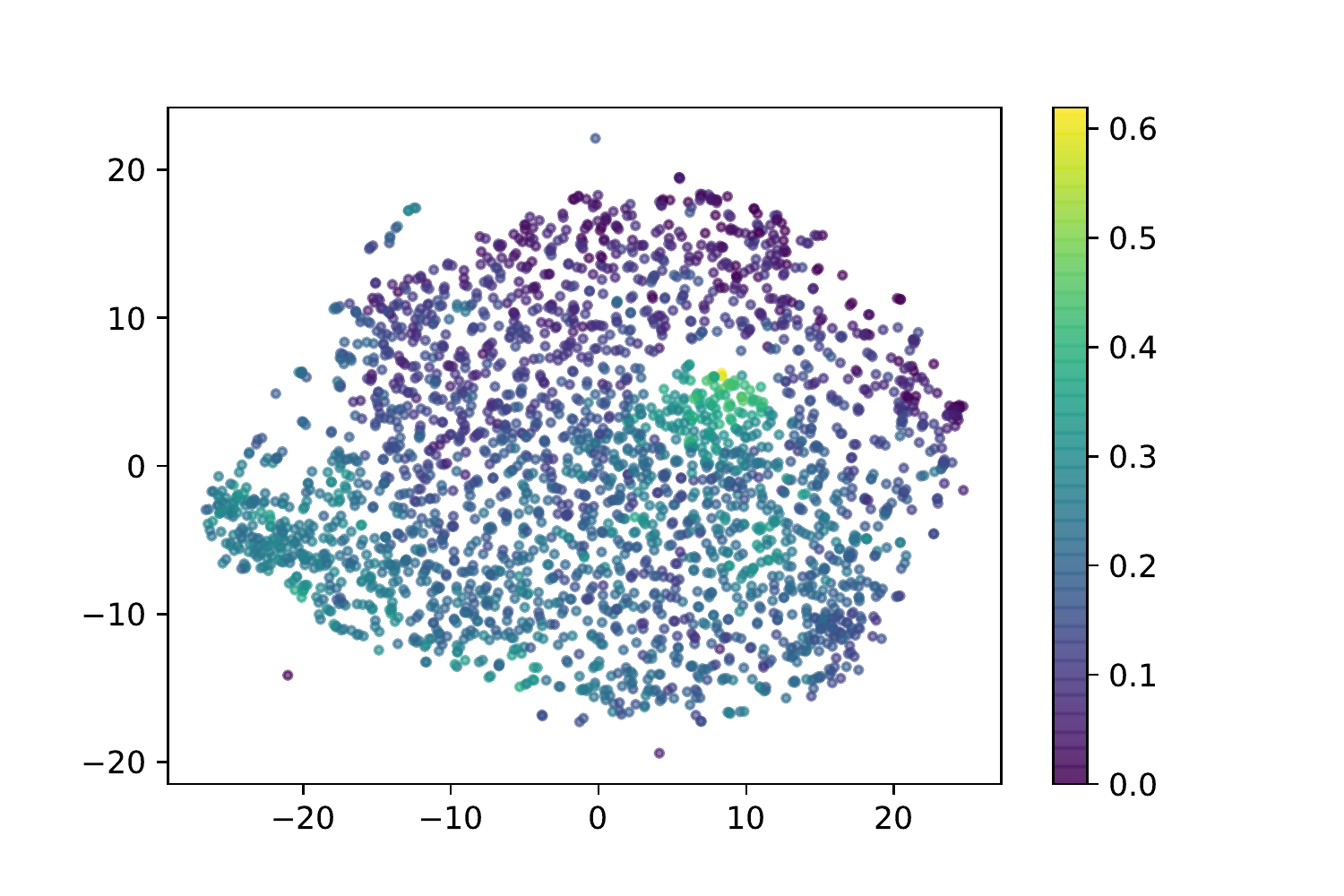}
	\includegraphics[width=.45\textwidth]{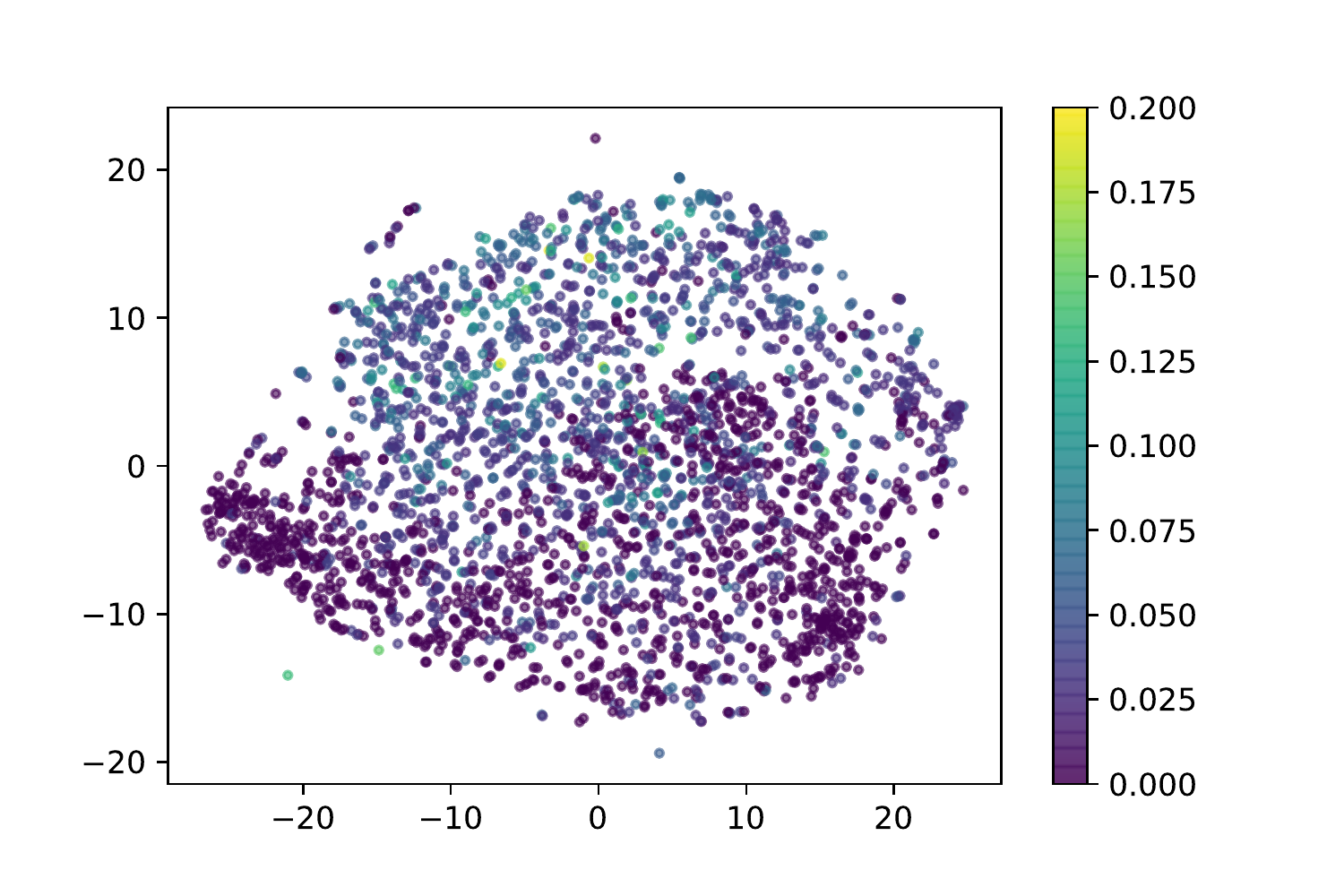}
	\includegraphics[width=.45\textwidth]{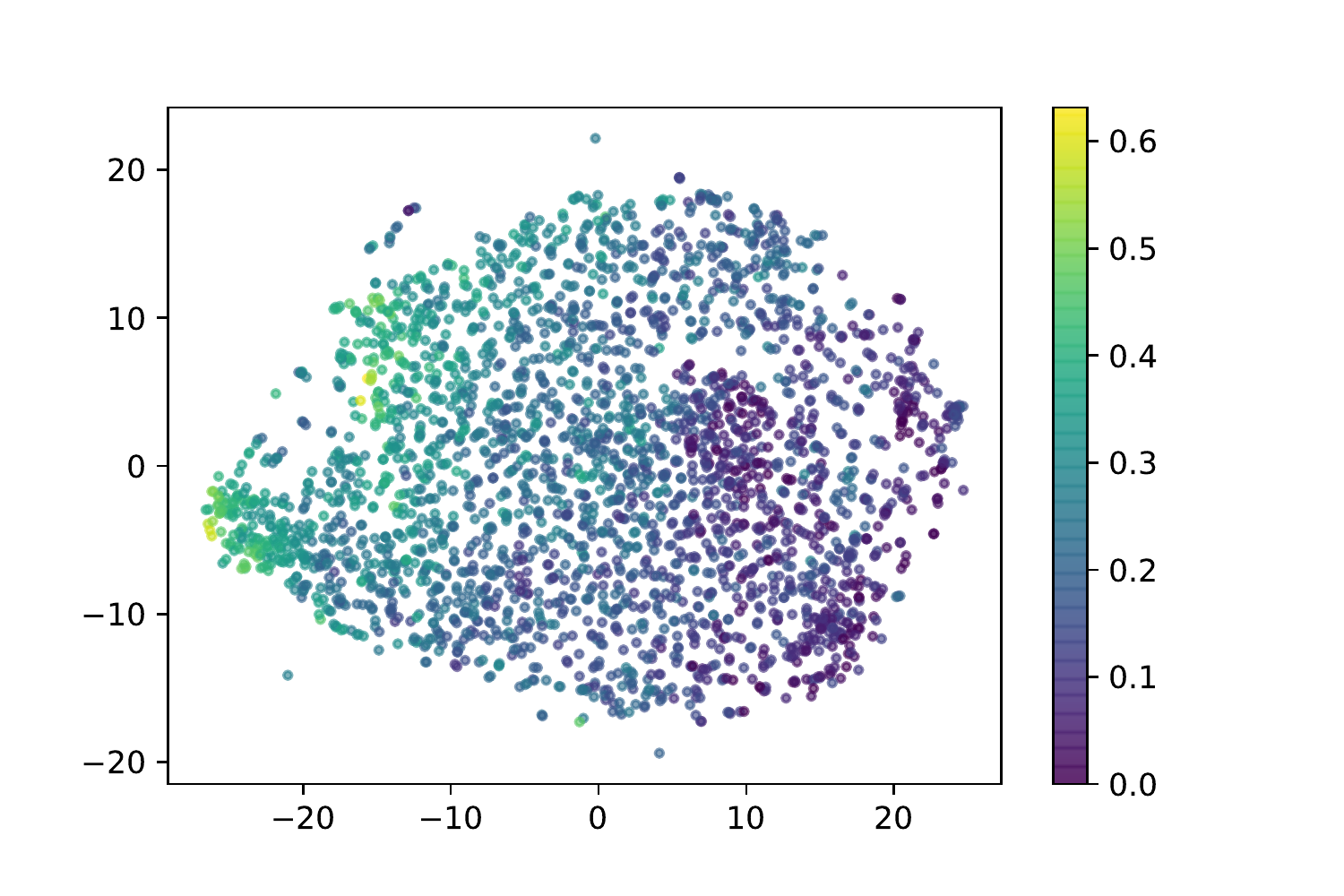}
	\includegraphics[width=.45\textwidth]{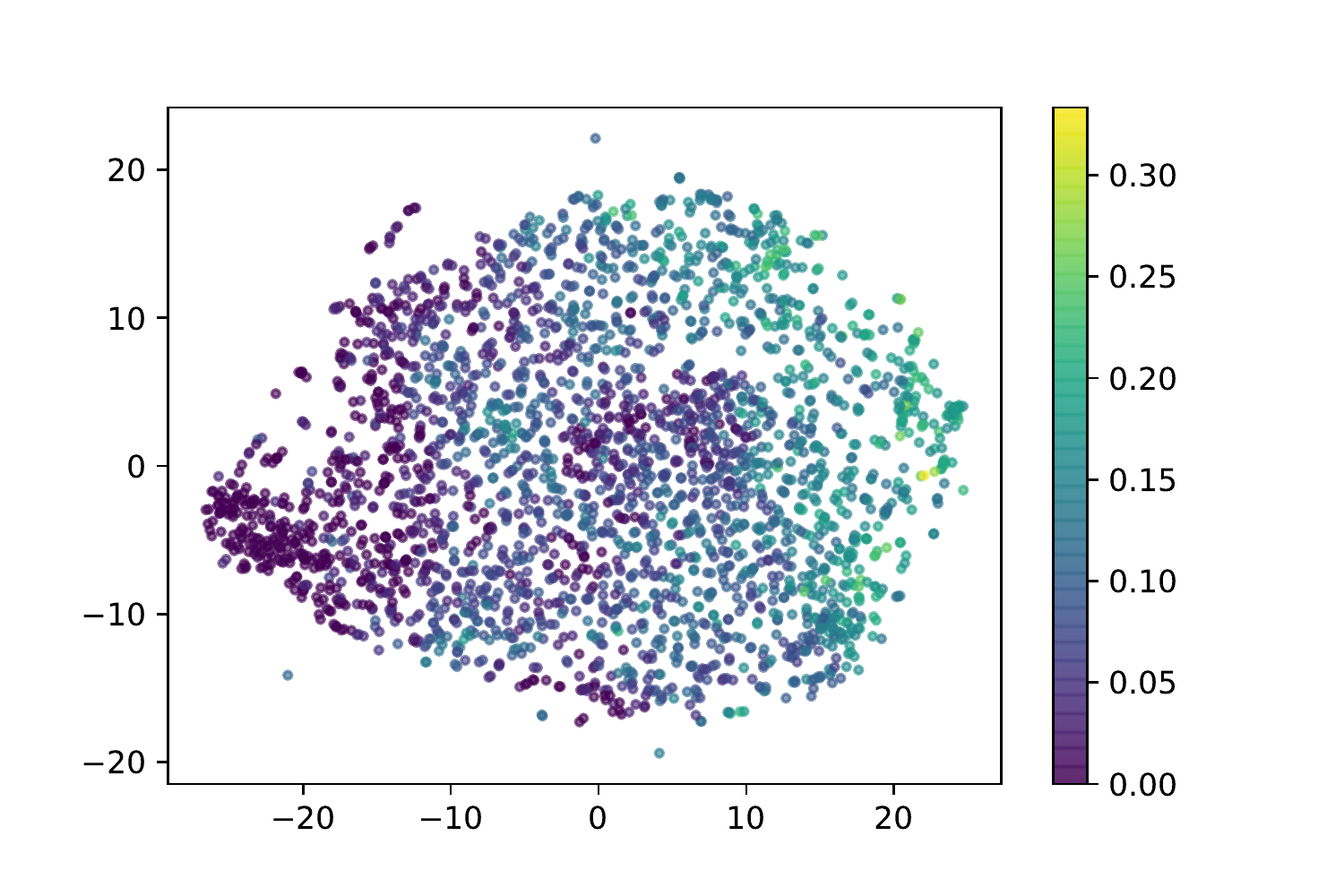}
	\includegraphics[width=.45\textwidth]{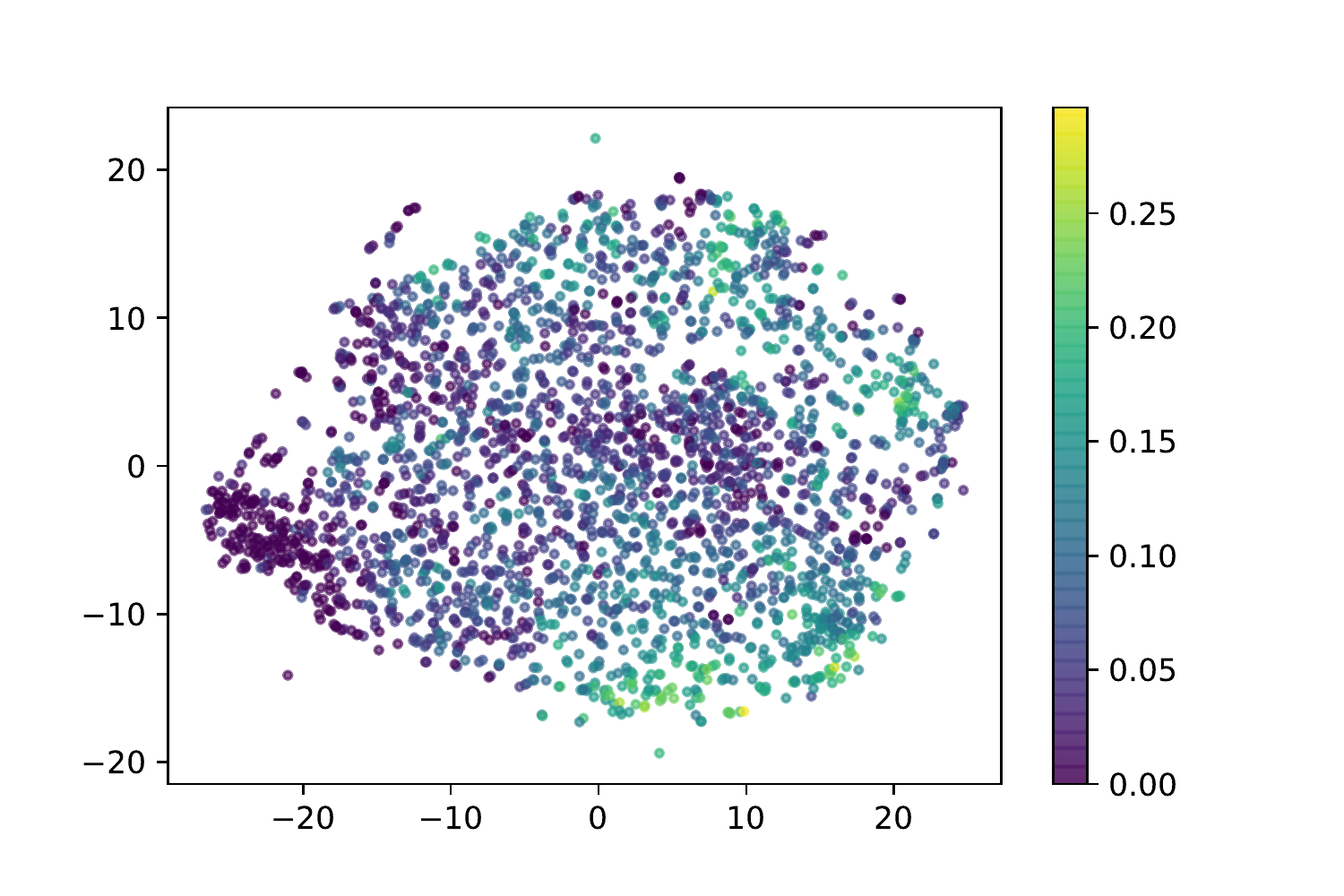}
	\includegraphics[width=.45\textwidth]{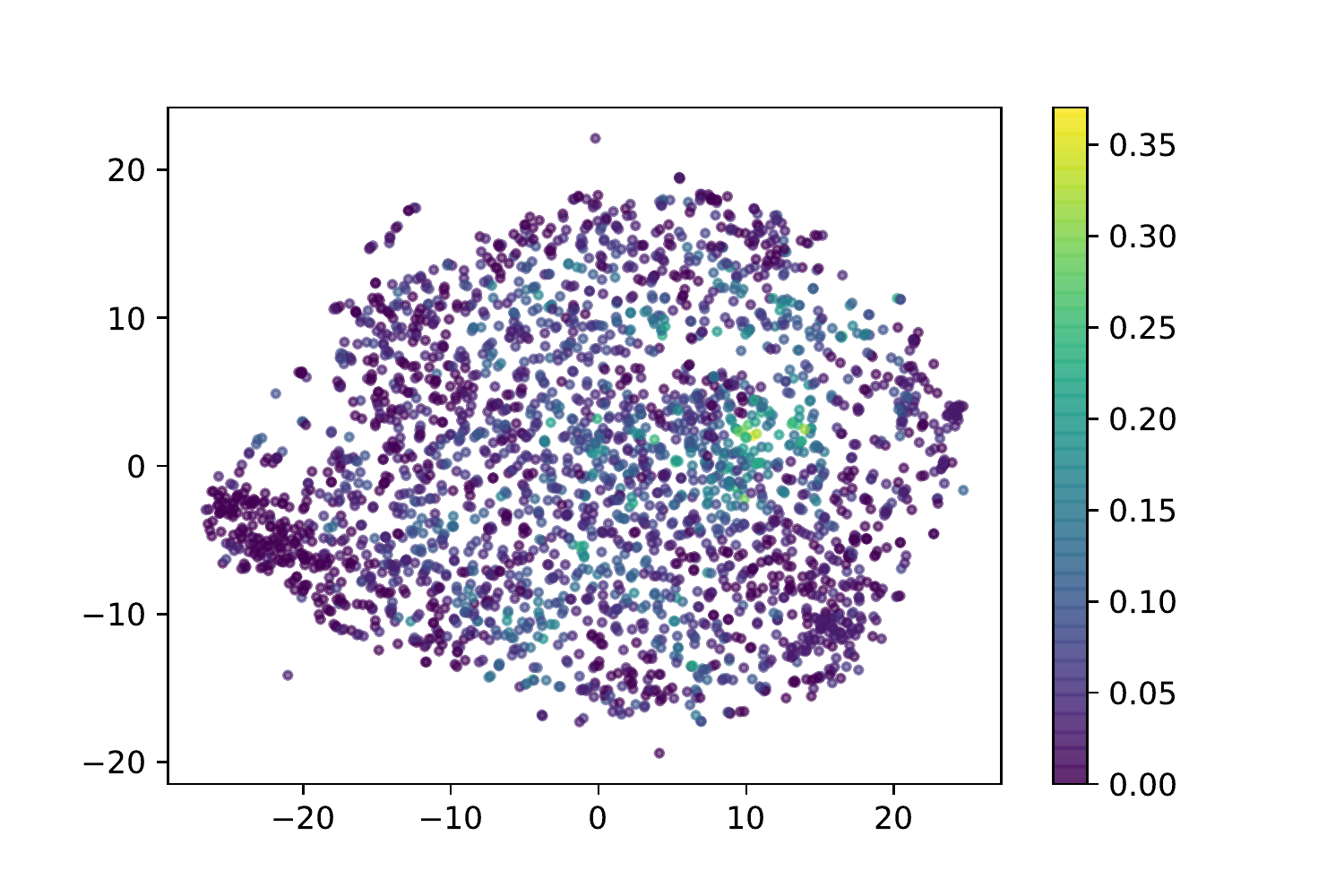}
\caption{Normalized feature embedding, where color represents the proportion of various cards in a player's deck at end of the game. Top-left is \emph{Copper}, top-right is \emph{Mines}, middle-left is \emph{Silvers} and \emph{Golds}, middle-right is \emph{Villages}, bottom-left is \emph{Markets}, and bottom-right is \emph{Militias} and \emph{Moats} .}
\label{main}
\end{center}
\end{figure*}

We see that the clustering isn't dominated by trace length, see Figure \ref{prop}. We further note that opening moves do not dominate the clustering, despite having the highest marginal effect on deck proportions and proliferating throughout the game state for all subsequent turns. For example, consider the popular starting card \emph{Village} in Figure \ref{village}.

\begin{figure*}
\begin{center}
	\includegraphics[width=.45\textwidth]{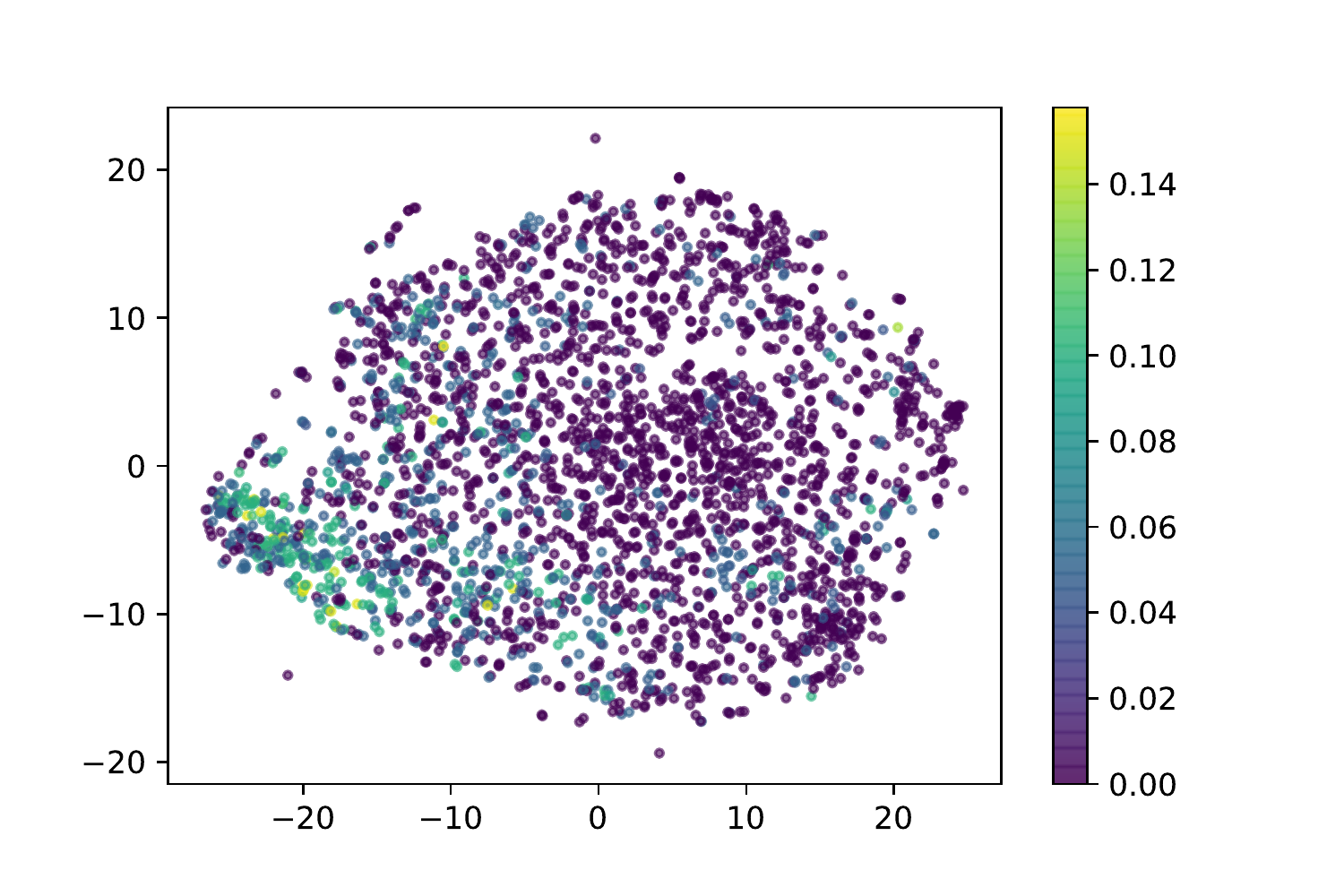}
	\includegraphics[width=.45\textwidth]{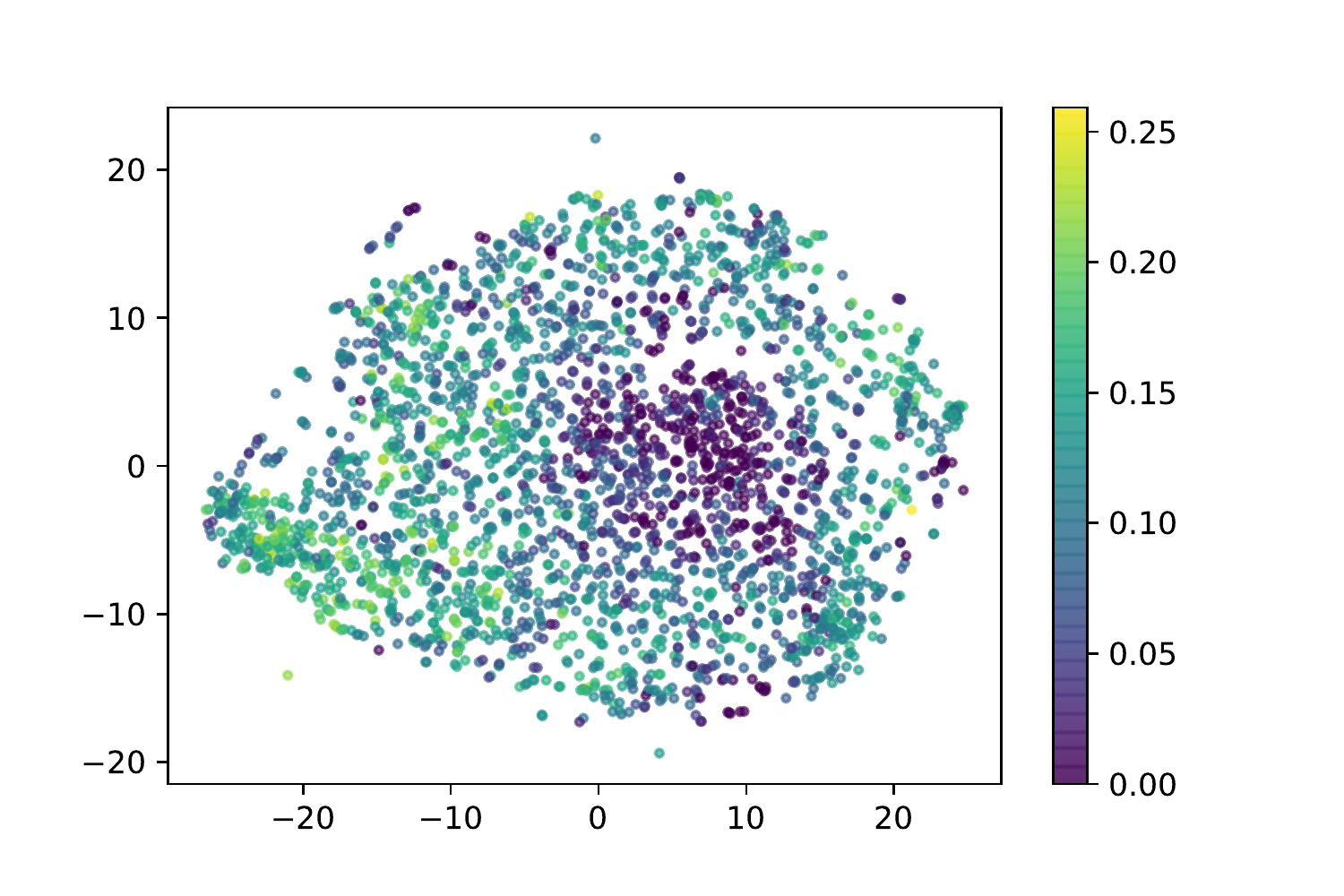}
\caption{Normalized feature embedding. On the left, color represents the proportion of \emph{Provinces} in the player's deck on turn 10. On the right, color represents the proportion of \emph{Provinces} at in the player's deck at the end of the game.}
\label{provinces}
\end{center}
\end{figure*}

However, this encoding does have a very interesting interpretation. The central cluster contains decks that have a high density of \emph{Coppers} at the end of the game (see Figure \ref{main}). \emph{Coppers} are the starting cards, and are very weak. Some of these traces simply did nothing on their turns, or performed poorly. This cluster represents little movement from the starting state, and the ring around it represents different ways of diverging from that state.

To the top of the t-SNE embedding, we see decks that rely heavily on \emph{Mines}, a card that lets a player replace their low-quality \emph{Coppers} with \emph{Silvers} and \emph{Golds}. To the left, we see so-called ``Big Money" players, going for currency. On the right, we see players who bought a lot of \emph{Villages}, a requirement for an action-heavy deck. Other action cards, like \emph{Smithy}, which synergizes well with \emph{Village}, are also dense on the right. Finally, we see a \emph{Market} strategy on the bottom.

Because we are not encoding plays, we cannot confirm how these synergies are being utilized, but we can apply conventional wisdom to the apparent correlation in the purchase of multiple cards. These figures reflect intuition about this cardset, that a player will either want to focus on action cards, which requires \emph{Villages} to be able to string those actions together, or focus on buying high-quality currency cards. These are the left and right hemispheres of the embedding.

We also see \emph{Militia} and \emph{Moat} concentrated in the center of the embedding. \emph{Militia} is a combative card that hurts other players and \emph{Moats} defend against these attacks, so it matches our expectation that these cards are associated with not moving as far from the starting state.

In the previous, non-proportional representations of decks, game length has a disproportional effect on encoding, because decks naturally get bigger. However, by looking at proportions, we can see the primary goal of the game is to start with the weak starting deck and produce something that has a high concentration of strong cards.

\emph{Provinces} are the main game-winning card, worth the most victory points. In Figure \ref{provinces}, we can see that as expected the players who have traveled the most distance from the starting deck are the ones with \emph{Provinces}. We note that at turn 10, the players with the highest concentration of \emph{Provinces} were those in the Big Money camp. Because of \emph{Dominion's} nature as an engine building game, different strategies get their engines functional at different rates, and this indicates that Big Money is a better strategy for shorter games. Other strategies start buying \emph{Provinces} later, but might be able to buy them more consistently, making them better late-game strategies.

\section{Future Work}
This work proposed a method for representing stochastic, varied-length games like \emph{Dominion}, by encoding game features as proportions. In the more general case, this method can be extended to other games by encoding the relative presence of game features in each game state, thus controlling for the divergence of the game state vector.

The method used above, where end-game states are resampled to fill out empty features, has the advantage that it allows early turns, which are the most directly comparable turns, to be compared directly. The next step towards a more robust model would be to approximate game stage, and resample intermediate game states from shorter games to line up stages of the game. This would better allow us to talk about distance between both players' movements across the surface of the game space, comparing the early game of one trace to the early game of the other, and late game to late game. This would have to be more complex than sampling proportionally, because the early turns are directly comparable agnostic to game length, meaning the later turns would need to be resampled more to compensate.

Another important further step is to encode cards by their effects, so as to allow comparison of strategies for games that used different card sets. We might still observe the dichotomy of currency cards versus action-enabling cards. In that situation we could talk about the ways in which cards serve similar roles in different sets, and imagine the player who would want to buy them. 

\bibliographystyle{aaai}
\bibliography{thesis}

\end{document}